\documentclass[11pt,a4paper]{article}
\usepackage[nohyperref]{acl2020}
\usepackage{times}
\usepackage{latexsym}
\usepackage{amssymb}
\usepackage{amsmath}
\usepackage{bm}
\usepackage{booktabs}

\usepackage{url}
\usepackage{tikz}
\usetikzlibrary{positioning, backgrounds, fit, arrows}

\usepackage{subcaption}
\usepackage{lmodern}

\usepackage{microtype}

\aclfinalcopy 

\title{Decoupling entrainment from consistency using deep neural networks}

\author{Andreas Weise \\
  Dept. of Computer Science \\
  The Graduate Center, CUNY \\
  365 5th Ave, New York, NY 10016 \\
  {\tt aweise@gradcenter.cuny.edu} \\\And
  Rivka Levitan \\
  Dept. of Computer and Information Science \\
  Brooklyn College, CUNY \\
  2900 Bedford Ave, Brooklyn, NY 11210 \\
  {\tt rlevitan@brooklyn.cuny.edu} \\}

\date{}

\begin{document}
\maketitle
\begin{abstract}
  Human interlocutors tend to engage in adaptive behavior known as \textit{entrainment} to become more similar to each other. Isolating the effect of consistency, i.e., speakers adhering to their individual styles, is a critical part of the analysis of entrainment. We propose to treat speakers' initial vocal features as confounds for the prediction of subsequent outputs. Using two existing neural approaches to deconfounding, we define new measures of entrainment that control for consistency. These successfully discriminate real interactions from fake ones. Interestingly, our stricter methods correlate with social variables in opposite direction from previous measures that do not account for consistency. These results demonstrate the advantages of using neural networks to model entrainment, and raise questions regarding how to interpret prior associations of conversation quality with entrainment measures that do not account for consistency.
\end{abstract}

\section{Introduction}

\textit{Entrainment} is a well-known psycholinguistic phenomenon causing people to adapt to conversation partners so as to become more similar. It affects many linguistic features including phonetics \cite{Pardo2006}, lexical choice \cite{Niederhoffer2002}, syntax \cite{Reitter2006}, and prosody \cite{Levitan2011a}. Importantly, it correlates with interesting aspects of the conversation such as task success, liking, and even rapport with a robot \cite{Nenkova2008,Ireland2011,Lubold2015}.

The researchers cited above employed various means to measure entrainment, such as correlations, models of conditional probabilities, comparisons of distributions, and perceived similarity. Recently, \citet{Nasir2018} proposed the first neural entrainment measure. Our work builds on theirs by addressing a challenge critical to measuring entrainment: accounting for consistency. 

Entrainment is defined as an active, though unconscious, adaptation of a speaker towards their partner. In practice, however, the static similarity or correlation between two speakers is often measured. Thus, even two speakers whose vocal characteristics were initially similar are perceived to have entrained, although no adaptation has taken place. Alternatively, when Speaker B entrains to Speaker A, both speakers are perceived to have entrained, without adaptation from Speaker A. We apply neural methods proposed by \citet{Pryzant2018} to explicitly deconfound \textit{consistency}, the tendency to adhere to one's own vocal style, from \textit{entrainment}, the tendency to adapt to one's partner. We argue that entrainment measures that do not control for \textit{consistency} overestimate the degree of entrainment in a conversation.

Section \ref{sec:data} explains the data and features that we use to train our networks, which are described in Section \ref{sec:methods}. Section \ref{sec:experiments} introduces two experiments to validate our methods whose results are discussed, lastly, in Section \ref{sec:discussion}.

\section{Data}\label{sec:data}

\subsection{Corpora}

To train our neural networks and for part of our experiments, we use the Fisher Corpus \cite{Cieri2004}, which consists of 11699 transcribed phone conversations in English. In each session, two previously unacquainted subjects discuss a given topic for ten minutes. We split the corpus into first 80\% for training, next 10\% for validation, and last 10\% for testing.

For additional experimentation, we use the Objects Games portion of the Columbia Games Corpus \cite{Benus2007}. It contains 12 dyadic, in-person, task-oriented conversations (sessions), about four hours of speech, which were fully transcribed. Each pair of speakers performed 14 tasks and each of the 168 tasks was annotated by five crowdworkers with social variables such as ``Is Person A encouraging their partner?'' \cite{Gravano2011}.

We use the transcriptions for both corpora to extract turns, i.e., sequences of speech from a single speaker without interruption by the interlocutor. Each turn consists of one or more transcription segments which exclude long pauses. We refer to these segments as inter-pausal units, IPUs, and use them as the basis of our analysis.

\subsection{Features}

For each turn-final or turn-initial IPU we extract 228 features in three steps, using a subset of the features for the INTERSPEECH 2010 Paralinguistic Challenge \cite{Schuller2010}.

First, we use openSMILE \cite[v2.3.0]{Eyben2013} to extract 38 low-level descriptors (LLDs). Smoothed pitch and its first-order delta as well as shimmer and two types of jitter are extracted using Gaussian windows (width=60ms, step=10ms, $\sigma$=0.25). Pitch smoothing is done with a median-filter of window size 5 frames to mitigate halving and doubling errors. Loudness and its first-order delta as well as 15 Mel-frequency cepstral coefficients, 8 Mel-frequency bands, and 8 line spectral pair frequencies are extracted using Hamming windows (width=25ms, step=10ms). 

Second, all LLDs are z-score normalized per speaker and session: $z = (x - \mu) / \sigma$ where $x$ is the original value and $\mu$ and $\sigma$ are the mean and standard deviation, respectively, of that LLD across all frames in all turn-initial and turn-final IPUs of the same speaker and session. 

Finally, openSMILE is used again to apply six functionals to each of the 38 sequences of normalized LLDs: mean, median, standard deviation, 1st percentile, 99th percentile, and the range between 1st and 99th percentile. This results in a total of 228 features per IPU.

\subsection{IPU triplets}

Each data sample we use consists of three IPUs. Two of them form a turn exchange, one being the final IPU of a turn from one speaker, the other being the first IPU of the next turn from the interlocutor. We refer to the turn-final IPU in a sample as \emph{IPU1}, the immediately following turn-initial IPU as \emph{IPU2} and their feature vectors as $\bm{x}_1$ and $\bm{x}_2$, respectively. Note that either speaker in a session represents \emph{IPU1} for some samples and \emph{IPU2} for others.  

The third IPU in each sample is the very first one uttered by the same speaker and in the same session as \emph{IPU2}. We refer to it as \emph{IPU0} and its feature vector as $\bm{x}_0$. Adding it to each sample is intended to help isolate the entrainment effect that \emph{IPU1} has on the production of \emph{IPU2}. We use the very first IPU for this purpose rather than, say, the most recent turn-initial IPU before \emph{IPU2} since such an IPU would itself be partially a result of entrainment to the interlocutor.

\begin{figure*}
  \centering
  \begin{subfigure}[t]{0.48\textwidth}
    \centering
    \begin{tikzpicture}
      \node[minimum width=0.5cm, minimum height=2cm, label=center:$\bm{x}_0$] (x0) {};
      \node[right=1mm of x0, minimum width=1cm, minimum height=2cm, label={[font=\small]center:ENC0}] (enc0) {};
      \node[right=1mm of enc0, minimum width=0.25cm, minimum height=1cm] (e0) {};
      \node[right=1mm of e0, minimum width=1cm, minimum height=2cm, label={[font=\small]center:DEC0}] (dec0) {};
      \node[right=1mm of dec0, minimum width=0.5cm, minimum height=2cm, label=center:$\hat{\bm{x}}_2^{\prime}$] (o1) {};
      \node[above=1mm of o1, minimum width=0.5cm, minimum height=2cm, label=center:$\bm{x}_1$] (x1) {};
      \node[right=1mm of o1, minimum width=1cm, minimum height=2cm] (dummy) {};
      \node[above=-20.1mm of dummy, minimum width=1cm, minimum height=4.1cm, label={[font=\small]center:ENC1}] (enc1) {};
      \node[right=1mm of enc1, minimum width=0.25cm, minimum height=1cm] (e1) {};
      \node[right=1mm of e1, minimum width=1cm, minimum height=2cm, label={[font=\small]center:DEC1}] (dec1) {};
      \node[right=1mm of dec1, minimum width=0.5cm, minimum height=2cm, label=center:$\hat{\bm{x}}_2$] (o2) {};

      \begin{scope}[on background layer]
	\fill[black] (x0.north west)--(x0.north east)--(x0.south east)--(x0.south west)--cycle;
	\fill[red!30] ([xshift=0.2mm,yshift=-0.2mm]x0.north west)--([xshift=-0.2mm,yshift=-0.2mm]x0.north east)--([xshift=-0.2mm,yshift=0.2mm]x0.south east)--([xshift=0.2mm,yshift=0.2mm]x0.south west)--cycle;
	\fill[black] (enc0.north west)--([yshift=-5mm]enc0.north east)--([yshift=5mm]enc0.south east)--(enc0.south west)--cycle;
	\fill[gray!30] ([xshift=0.2mm,yshift=-0.3mm]enc0.north west)--([xshift=-0.2mm,yshift=-5.2mm]enc0.north east)--([xshift=-0.2mm,yshift=5.2mm]enc0.south east)--([xshift=0.2mm,yshift=0.3mm]enc0.south west)--cycle;
	\fill[black] (e0.north west)--(e0.north east)--(e0.south east)--(e0.south west)--cycle;
	\fill[red!30] ([xshift=0.2mm,yshift=-0.2mm]e0.north west)--([xshift=-0.2mm,yshift=-0.2mm]e0.north east)--([xshift=-0.2mm,yshift=0.2mm]e0.south east)--([xshift=0.2mm,yshift=0.2mm]e0.south west)--cycle;
	\fill[black] ([yshift=-5mm]dec0.north west)--(dec0.north east)--(dec0.south east)--([yshift=5mm]dec0.south west)--cycle;
	\fill[gray!30] ([xshift=0.2mm,yshift=-5.2mm]dec0.north west)--([xshift=-0.2mm,yshift=-0.3mm]dec0.north east)--([xshift=-0.2mm,yshift=0.3mm]dec0.south east)--([xshift=0.2mm,yshift=5.2mm]dec0.south west)--cycle;
	\fill[black] (o1.north west)--(o1.north east)--(o1.south east)--(o1.south west)--cycle;
	\fill[red!30] ([xshift=0.2mm,yshift=-0.2mm]o1.north west)--([xshift=-0.2mm,yshift=-0.2mm]o1.north east)--([xshift=-0.2mm,yshift=0.2mm]o1.south east)--([xshift=0.2mm,yshift=0.2mm]o1.south west)--cycle;
	\fill[black] (x1.north west)--(x1.north east)--(x1.south east)--(x1.south west)--cycle;
	\fill[red!30] ([xshift=0.2mm,yshift=-0.2mm]x1.north west)--([xshift=-0.2mm,yshift=-0.2mm]x1.north east)--([xshift=-0.2mm,yshift=0.2mm]x1.south east)--([xshift=0.2mm,yshift=0.2mm]x1.south west)--cycle;
	\fill[black] (enc1.north west)--([yshift=-15.5mm]enc1.north east)--([yshift=15.5mm]enc1.south east)--(enc1.south west)--cycle;
	\fill[gray!30] ([xshift=0.2mm,yshift=-0.5mm]enc1.north west)--([xshift=-0.2mm,yshift=-15.7mm]enc1.north east)--([xshift=-0.2mm,yshift=15.7mm]enc1.south east)--([xshift=0.2mm,yshift=0.5mm]enc1.south west)--cycle;
	\fill[black] (e1.north west)--(e1.north east)--(e1.south east)--(e1.south west)--cycle;
	\fill[red!30] ([xshift=0.2mm,yshift=-0.2mm]e1.north west)--([xshift=-0.2mm,yshift=-0.2mm]e1.north east)--([xshift=-0.2mm,yshift=0.2mm]e1.south east)--([xshift=0.2mm,yshift=0.2mm]e1.south west)--cycle;
	\fill[black] ([yshift=-5mm]dec1.north west)--(dec1.north east)--(dec1.south east)--([yshift=5mm]dec1.south west)--cycle;
	\fill[gray!30] ([xshift=0.2mm,yshift=-5.2mm]dec1.north west)--([xshift=-0.2mm,yshift=-0.3mm]dec1.north east)--([xshift=-0.2mm,yshift=0.3mm]dec1.south east)--([xshift=0.2mm,yshift=5.2mm]dec1.south west)--cycle;
	\fill[black] (o2.north west)--(o2.north east)--(o2.south east)--(o2.south west)--cycle;
	\fill[red!30] ([xshift=0.2mm,yshift=-0.2mm]o2.north west)--([xshift=-0.2mm,yshift=-0.2mm]o2.north east)--([xshift=-0.2mm,yshift=0.2mm]o2.south east)--([xshift=0.2mm,yshift=0.2mm]o2.south west)--cycle;
      \end{scope}
    \end{tikzpicture}
    \caption{Deep Residualization network for measure $\bf{DR}$.}\label{fig:net_DR}
  \end{subfigure}
  ~
  \begin{subfigure}[t]{0.48\textwidth}
    \centering
    \begin{tikzpicture}
      \node[minimum width=0.5cm, minimum height=2cm, label=center:$\bm{x}_1$] (x1) {};
      \node[right=1mm of x1, minimum width=1cm, minimum height=2cm, font=\sffamily\scriptsize, label={[font=\small]center:ENC0}] (enc0) {};
      \node[right=1mm of enc0, minimum width=0.5cm, minimum height=1cm, label=center:$\bm{e}$] (e0) {};
      \node[above=0.3mm of e0, minimum width=0.5cm, minimum height=1cm] (dummy1) {};
      \node[below=0.3mm of e0, minimum width=0.5cm, minimum height=1cm] (dummy2) {};
      \node[right=22mm of dummy1, minimum width=1cm, minimum height=2cm, font=\sffamily\scriptsize, label={[font=\small]center:DEC1}] (dec0) {};
      \node[right=22mm of dummy2, minimum width=1cm, minimum height=2cm, font=\sffamily\scriptsize, label={[font=\small]center:DEC2}] (dec1) {};
      \node[right=1mm of dec0, minimum width=0.5cm, minimum height=2cm, label=center:$\hat{\bm{x}}_2$] (x2) {};
      \node[right=1mm of dec1, minimum width=0.5cm, minimum height=2cm, label=center:$\hat{\bm{x}}_0$] (x0) {};
      \node[left=5mm of dec1, minimum width=0.1cm, minimum height=0.5cm] (gr) {};
      \node[align=left, below=0mm of gr, font=\sffamily\scriptsize] {reversal\\ layer};

      \begin{scope}[on background layer]
	\draw[thick,->] ([xshift=1mm]e0.east) -- +(0.5,0) -- +(1,1.03) -> +(2,1.03);
	\draw[thick,->] ([xshift=1mm]e0.east) -- +(0.5,0) -- +(1,-1.03) -> +(2,-1.03);
	\fill[black] (x1.north west)--(x1.north east)--(x1.south east)--(x1.south west)--cycle;
	\fill[red!30] ([xshift=0.2mm,yshift=-0.2mm]x1.north west)--([xshift=-0.2mm,yshift=-0.2mm]x1.north east)--([xshift=-0.2mm,yshift=0.2mm]x1.south east)--([xshift=0.2mm,yshift=0.2mm]x1.south west)--cycle;
	\fill[black] (enc0.north west)--([yshift=-5mm]enc0.north east)--([yshift=5mm]enc0.south east)--(enc0.south west)--cycle;
	\fill[gray!30] ([xshift=0.2mm,yshift=-0.3mm]enc0.north west)--([xshift=-0.2mm,yshift=-5.2mm]enc0.north east)--([xshift=-0.2mm,yshift=5.2mm]enc0.south east)--([xshift=0.2mm,yshift=0.3mm]enc0.south west)--cycle;
	\fill[black] (e0.north west)--(e0.north east)--(e0.south east)--(e0.south west)--cycle;
	\fill[red!30] ([xshift=0.2mm,yshift=-0.2mm]e0.north west)--([xshift=-0.2mm,yshift=-0.2mm]e0.north east)--([xshift=-0.2mm,yshift=0.2mm]e0.south east)--([xshift=0.2mm,yshift=0.2mm]e0.south west)--cycle;
	\fill[black] ([yshift=-5mm]dec0.north west)--(dec0.north east)--(dec0.south east)--([yshift=5mm]dec0.south west)--cycle;
	\fill[gray!30] ([xshift=0.2mm,yshift=-5.2mm]dec0.north west)--([xshift=-0.2mm,yshift=-0.3mm]dec0.north east)--([xshift=-0.2mm,yshift=0.3mm]dec0.south east)--([xshift=0.2mm,yshift=5.2mm]dec0.south west)--cycle;
	\fill[black] ([yshift=-5mm]dec1.north west)--(dec1.north east)--(dec1.south east)--([yshift=5mm]dec1.south west)--cycle;
	\fill[gray!30] ([xshift=0.2mm,yshift=-5.2mm]dec1.north west)--([xshift=-0.2mm,yshift=-0.3mm]dec1.north east)--([xshift=-0.2mm,yshift=0.3mm]dec1.south east)--([xshift=0.2mm,yshift=5.2mm]dec1.south west)--cycle;
	\fill[black] (x2.north west)--(x2.north east)--(x2.south east)--(x2.south west)--cycle;
	\fill[red!30] ([xshift=0.2mm,yshift=-0.2mm]x2.north west)--([xshift=-0.2mm,yshift=-0.2mm]x2.north east)--([xshift=-0.2mm,yshift=0.2mm]x2.south east)--([xshift=0.2mm,yshift=0.2mm]x2.south west)--cycle;
	\fill[black] (x0.north west)--(x0.north east)--(x0.south east)--(x0.south west)--cycle;
	\fill[red!30] ([xshift=0.2mm,yshift=-0.2mm]x0.north west)--([xshift=-0.2mm,yshift=-0.2mm]x0.north east)--([xshift=-0.2mm,yshift=0.2mm]x0.south east)--([xshift=0.2mm,yshift=0.2mm]x0.south west)--cycle;
	\fill[black] (gr.north west)--(gr.north east)--(gr.south east)--(gr.south west)--cycle;
	\fill[red] ([xshift=0.2mm,yshift=-0.2mm]gr.north west)--([xshift=-0.2mm,yshift=-0.2mm]gr.north east)--([xshift=-0.2mm,yshift=0.2mm]gr.south east)--([xshift=0.2mm,yshift=0.2mm]gr.south west)--cycle;
      \end{scope}

    \end{tikzpicture}
    \caption{Adversarial network for measure $\bf{A}$.}\label{fig:net_A}
  \end{subfigure}
  \caption{Two network architectures used to define new entrainment measures. Encoders and decoders are displayed as gray trapezoids; input, output, and intermediate vectors as light red rectangles.}
\label{fig:nets}
\end{figure*}
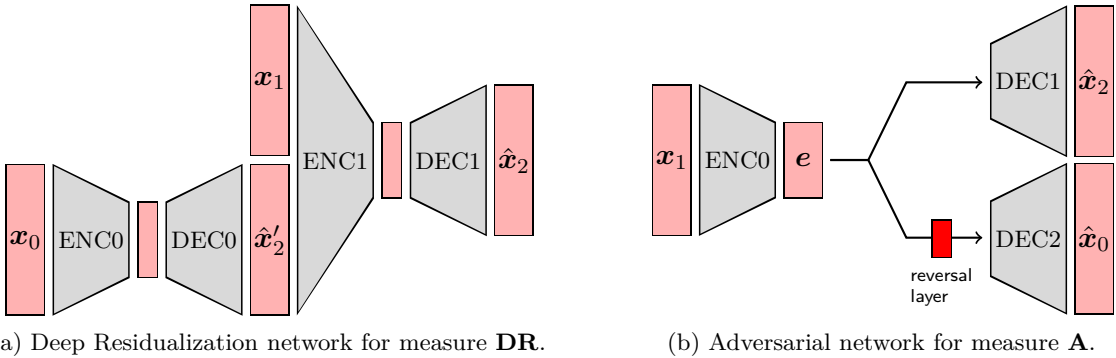

\section{Neural Entrainment Measures}\label{sec:methods}

In order to isolate the entrainment effect that the turn-final \emph{IPU1} has on the production of the subsequent turn-initial \emph{IPU2} of the interlocutor, we apply two neural network architectures (see Figure \ref{fig:nets}) based on deconfounding methods proposed by \citet{Pryzant2018}. 

\subsection{Deep Residualization Measure $\bf{DR}$}

Our first neural entrainment measure determines how much the prediction of $\bm{x}_2$ from $\bm{x}_0$ improves through the additional input of $\bm{x}_1$. That is, we first train a subnetwork to predict $\bm{x}_2$ from $\bm{x}_0$ as well as possible. This is meant to capture consistency, how much the speaker maintains their initial speaking style. We then freeze that subnetwork's parameters and train a second part of the overall network. This part takes the concatenation of the intermediate prediction $\hat{\bm{x}_2}^\prime$ and $\bm{x}_1$ as its input to produce a final prediction $\hat{\bm{x}_2}$. 

Figure \ref{fig:net_DR} illustrates this Deep Residualization network. It consists of two encoders ENC0 and ENC1 and two decoders DEC0 and DEC1, feedforward neural networks that produce under- and over-representations of their input, respectively. For the training of both subnetworks, we use the whole training set in mini-batches of 128 samples, the Adam optimizer \cite{Kingma2014}, and a loss function based on the smooth L1 norm, which is defined as follows:
\begin{align}
  \mathcal{L}(\bm{x},\bm{y}) & = \sum_{i=1}^{N} \text{smooth}_{L1}(x_i - y_i)  \label{eq:loss}\\
  \text{smooth}_{L1}(d) & = 
  \begin{cases} 
    0.5d^2, & \text{if}\ |d| < 1 \\
    |d| - 0.5, & \text{otherwise} 
  \end{cases}
\end{align}

We experiment with different component widths and depths but find that they do not significantly affect performance on the validation set. Therefore, we adopt the architecture of \citet{Nasir2018} for direct comparison. Each component has two fully connected layers. The output of the first layer is batch normalized, processed by Rectified Linear Units (ReLU), and then passed on to the second layer: 
\begin{align*}
  \text{ENC}(\bm{x}) & = \bm{W}_2(ReLU(BN(\bm{W}_1\bm{x} + \bm{b}_1))) + \bm{b}_2\\
  \text{DEC}(\bm{e}) & = \bm{W}_4(ReLU(BN(\bm{W}_3\bm{e} + \bm{b}_3))) + \bm{b}_4
\end{align*}

The dimensions of $\bm{W}_1$, $\bm{W}_2$, $\bm{W}_3$, and $\bm{W}_4$ are $128\times228$, $30\times128$, $128\times30$, and $228\times128$, resp., except for ENC1 whose $\bm{W}_1$ has size $456\times128$. 

Using this network and the loss function defined in Equation \ref{eq:loss}, we define the following measure of entrainment for a turn exchange:
\begin{align*}
\bf{DR}(\bm{x}_0, \bm{x}_1, \bm{x}_2) & = \mathcal{L}(\bm{x}_2, \hat{\bm{x}}_2) - \mathcal{L}(\bm{x}_2, \hat{\bm{x}}_2^{\prime}), \\
\text{with}\hspace{25pt} \hat{\bm{x}_2}^{\prime} & = \text{DEC0}(\text{ENC0}(\bm{x}_0)) \\
\text{and}\hspace{31pt} \hat{\bm{x}_2} & = \text{DEC1}(\text{ENC1}(\bm{x}_1, \bm{x}_0)) 
\end{align*}

$\bf{DR}$ decreases as the final prediction (from the interlocutor's turn) improves on the intermediate prediction (from the speaker's own initial turn alone), i.e., as entrainment increases.

\subsection{Adversarial Measure $\bf{A}$}

For our second neural entrainment measure, we train a network (see Figure \ref{fig:net_A}) to produce an encoding $\bm{e}$ from $\bm{x}_1$ which is maximally predictive for $\bm{x}_2$ as well as minimally predictive for $\bm{x}_0$. The latter is achieved through a gradient reversal layer before the decoder for $\bm{x}_0$. This multiplies gradients by $-1$, encouraging encodings which maximize the loss for $\bm{x}_0$ while still producing a decoder that tries to minimize the same loss based on that encoding.

We use the same component design, the same training scheme, and the same loss function per decoder as for $\bf{DR}$. Based on this network we define the following entrainment measure: 
\begin{align*}
\bf{A}(\bm{x}_0, \bm{x}_1, \bm{x}_2) & = \mathcal{L}(\bm{x}_2, \hat{\bm{x}_2}) - \mathcal{L}(\bm{x}_0, \hat{\bm{x}}_0), \\
\text{with}\hspace{25pt} \hat{\bm{x}_2} & = \text{DEC1}(\text{ENC0}(\bm{x}_1)) \\
\text{and}\hspace{28pt} \hat{\bm{x}_0} & = \text{DEC2}(\text{ENC0}(\bm{x}_1)) 
\end{align*}

$\bf{A}$ becomes smaller for more accurate predictions of $\bm{x}_2$ and more inaccurate predictions of $\bm{x}_0$, that is, the more similarity (entrainment) exists between $\bm{x}_1$ and $\bm{x}_2$ independent of $\bm{x}_0$. Thus, it models the adaptation of $\bm{x}_2$ to $\bm{x}_1$, rather than the static similarity.

\section{Experiments}\label{sec:experiments}

To assess whether our entrainment measures capture useful information, we conduct two evaluations (following \citet{Nasir2018}).

\subsection{Recognizing fake sessions}\label{sec:fake_ses}

From each real session, we generate a fake one by shuffling $\bm{x}_1$ vectors across samples to create artificial turn exchanges. If the average of $\bf{DR}$ and $\bf{A}$, respectively, for all samples of a real session is lower than for the corresponding fake one, that measure is considered to have successfully identified the real session. This is because real turn exchanges should contain more entrainment than fake ones and both measures produce lower values for greater entrainment.

\begin{table}[t]\centering
\begin{tabular}{ccc}\toprule
 & \textbf{Fisher (test)} & \textbf{Games} \\
$\bf{DR}$ & 95.1\% (0.7\%) & 85.8\% (12.8\%) \\
$\bf{A}$  & 94.4\% (0.6\%) & 80.3\% (12.6\%) \\
\midrule
$\bf{NED}$ & 98.9\% (1.0\%) & --- \\
\bottomrule
\end{tabular} 
\caption{Accuracy of discriminating fake sessions from real ones (avg. and std. deviation for 30 runs)}\label{tab:exp1}
\end{table}

Table \ref{tab:exp1} shows the results of 30 runs for our measures and $\bf{NED}$ from \citet{Nasir2018} on the test set and, for our measures only, the Games Corpus. For each run, our networks are retrained and new fake sessions are created. We note that Nasir et al. may have used a different 10\% of the Fisher Corpus and apparently did not retrain their network for each run.

Both our measures are highly accurate for the Fisher Corpus and do well even for the Games Corpus on which the networks are not trained. Both perform slightly worse than $\bf{NED}$ but outperform all three baselines reported by Nasir et al. (best accuracy 92.3\%). 

\subsection{Correlations with social variables}

Each task in the Games Corpus is annotated with several social variables. We analyze three annotations for correlations with our entrainment measures (abbreviations highlighted): \emph{Is the speaker \textbf{enc}ouraging their partner? Trying to be \textbf{lik}ed? Trying to \textbf{dom}inate the conversation?} \citet{Levitan2012} considered these and found positive correlations between entrainment and the first two. For \textbf{\textit{dom}} they expected a positive correlation but found none.

For each social variable, we compute Pearson correlations between: 1) the number of annotators (out of five) who answered the question affirmatively for a given task and speaker and 2) the average entrainment, as per our measures, across all data samples from that task where that speaker responded, i.e., produced \emph{IPU0} and \emph{IPU2}. Unlike for Subsection \ref{sec:fake_ses}, we find that results vary greatly across the training runs of our networks\footnote{See Appendix \ref{app:pvals} for our standards of significance.} but with clear trends: $\bf{DR}$ correlates moderately with  \textbf{\textit{dom}}, weakly with \textbf{\textit{lik}} and not at all with  \textbf{\textit{enc}} while $\bf{A}$ does not exhibit any correlations. That is, \textbf{\textit{lik}} and \textbf{\textit{dom}} correlate \emph{negatively} with entrainment according to $\bf{DR}$, contrary to expectation and the results of \citet{Levitan2012}. Note that \textbf{\textit{lik}} and \textbf{\textit{dom}} are not correlated with each other ($r=0.04$, $p=0.44$). Table \ref{tab:exp2} lists the results for the $\bf{DR}$ network with the strongest correlation for \textbf{\textit{dom}} and the $\bf{A}$ network with the strongest correlation for \textbf{\textit{lik}}. 

\begin{table}[t]\centering
\begin{tabular}{lrrrr}\toprule
 & \multicolumn{2}{c}{$\bf{DR}$} & \multicolumn{2}{c}{$\bf{A}$} \\
 & $r$ & $p$ & $r$ & $p$ \\
\textbf{\textit{enc}} & -0.002 & 0.98 & -0.030 & 0.60 \\
\textbf{\textit{lik}} & \textbf{+0.22} & \textbf{8.9e-05} & +0.09 & 0.10 \\
\textbf{\textit{dom}} & \textbf{+0.39} & \textbf{7.5e-13} & +0.02 & 0.73 \\
\bottomrule
\end{tabular} 
\caption{Pearson correlations with social variables (significant results bold).}\label{tab:exp2}
\end{table}

\section{Discussion}\label{sec:discussion}

We propose two neural measures of entrainment that control for consistency. We empirically validate these measures by demonstrating their ability to discriminate between real and fake sessions. Although our measures perform slightly worse than the one reported by \citet{Nasir2018}, we believe this is because their measure captures both entrainment and consistency and therefore better describes the expected similarity between two turns, but is overly broad as a measure of entrainment.

Most intriguingly, the strict separation of consistency and entrainment leads to correlations that are very different from those with other entrainment measures that do not account for consistency, even on the same corpus. This resembles the results of \citet{Perez2016}, who found that correlations differ based on how \emph{dis}entrainment is treated. 

Our findings cast previous links between conversation quality and entrainment measures that do not account for consistency in a new light. It is worth revisiting those with the new ability to distinguish between consistency and entrainment.

In our future work, we intend to expand the network inputs for each prediction to the entire prior conversation context using RNNs with attention. We will also conduct further analysis of these entrainment measures, e.g., by feature, speaker sex, role, and dialogue act. 

\newpage

\bibliography{acl2020}
\bibliographystyle{acl_natbib}

\newpage

\appendix

\section{Multiple testing for correlations with social variables}\label{app:pvals}

The correlations between social variables and our entrainment measures vary greatly across retrainings of the underlying networks. This is especially true for $\bf{DR}$, with $p$-values for correlations with \textbf{\textit{dom}} ranging from 7.5e-13 to almost 1.

To address this, we retrained both networks 100 times, recomputing the Pearson correlations each time. To control the false discovery rate resulting from this multiple testing, we use the procedure of \citet{Benjamini1995}. Each run consists of three tests per measure. We sort each group of three tests by their $p$ values and determine the smallest value $\alpha$ such that $p_k \leq k * \alpha / 3$ for at least one $p$ value, where $k \in \{1,2,3\}$ is its position after sorting. Finally, we determine the largest $k$ such that $\alpha_k < k * 0.05 / 100$ where $\alpha_k$ is the $k$-th smallest $\alpha$ value for any run of the respective measure, the level at which at least one of three correlations is significant for that run and measure. 

Using this method, we find that 65 out of 100 times the correlation between $\bf{DR}$ and \textbf{\textit{dom}} is significant as well as 36 times for \textbf{\textit{lik}}. None of the correlations for $\bf{A}$ reach the level of significance, not even in terms of the ``raw'' $p$ values. 

For all but three of the 65 runs with significant correlations between $\bf{DR}$ and \textbf{\textit{dom}}, the correlation has the same valence. The three with opposite valence are among the weakest, the most significant one having only the 47th smallest $\alpha$ value. All 36 significant correlations between $\bf{DR}$ and \textbf{\textit{lik}} have the same valence. Considering the clear overall trends, we conclude that $\bf{DR}$ correlates positively with \textbf{\textit{dom}} and to a lesser degree with \textbf{\textit{lik}}.

\end{document}